\documentclass[twoside]{article}

\usepackage[utf8]{inputenc}

\title{Large-scale Analysis of Chess Games with Chess Engines: A Preliminary Report}
\author{Mathieu Acher and Fran\c{c}ois Esnault}

\usepackage{cite}
\usepackage{listings}
\usepackage{graphicx}
\usepackage{geometry}
\usepackage{subfig}
\usepackage{caption}

\usepackage{float}
\usepackage{url}
\usepackage{hyperref}

\hypersetup{
    pdftitle={Large-scale Analysis of Chess Games with Chess Engines: A Preliminary Report},    
    pdfauthor={Mathieu Acher and Francois Esnault},     
    pdfsubject={Chess},   
    pdfcreator={Mathieu Acher and Francois Esnault},   
    pdfproducer={Mathieu Acher and Francois Esnault}, 
    pdfkeywords={chess, data, artificial intelligence}, 
    linkcolor=red,          
    citecolor=green,        
    filecolor=magenta,      
    urlcolor=cyan           
}

\usepackage{RR}

\RRdate{April 2016}

\RRauthor{
Mathieu Acher\thanks[i1]{Inria/IRISA, University of Rennes 1, France}
\and Fran\c{c}ois Esnault\thanksref{i1}
}
\authorhead{Acher and Esnault}
\RRtitle{Analyse large échelle de parties d'échecs avec des moteurs: un rapport préliminaire} 
\RRetitle{Large-scale Analysis of Chess Games with Chess Engines: A Preliminary Report}
\titlehead{Large-scale Analysis of Chess Games with Chess Engines}
\RRresume{%
La force des moteurs d'échecs et l'existence de nombreuses parties d'échecs ont attiré l'attention des joueurs d'échecs, des scientifiques de la donnée,
et des chercheurs au cours des dernières années.
Les meilleurs moteurs fournissent un jugement autoritaire qui peut être utilisé dans de nombreuses applications comme
la détection de triches, le calcul d'une force intrinsèque, l'évaluation d'aptitudes, ou l'étude de prises de décision par des humains.
Un problème important pour la communauté de chercheurs est de collecter un large ensemble de parties d'échecs avec les jugements des moteurs.
Malheureusement l'analyse de chaque coup peut prendre énormément de temps.
Dans ce rapport, nous décrivons notre effort pour analyser près de 5 millions de parties d'échecs avec une grille de calcul.
Durant l'été 2015, nous avons analysé 270 millions de positions uniques issues de parties réelles en utilisant le moteur Stockfish avec une assez grande profondeur (20).
Nous avons construit une base de données d'évaluation d'échecs de plus de 1 tera-octet, représentant un temps estimé de 50 années sur une machine seule.
Notre travail est une première étape vers la réplication de résultats de recherche,
la mise à disposition de données ouvertes et de procédures pour explorer de nouvelles directions, et l'étude de problèmes de génie logiciel pour calculer des milliards de coups.
} 

\RRabstract{%
The strength of chess engines together with the availability of numerous chess games have attracted the attention of chess players, data scientists, and researchers during the last decades.
State-of-the-art engines now provide an authoritative judgement that can be used in many applications like cheating detection, intrinsic ratings computation, skill assessment, or the study of human decision-making.
 A key issue for the research community is to gather a large dataset of chess games together with the judgement of chess engines.
 Unfortunately the analysis of each move takes lots of times.
 In this paper, we report our effort to analyse almost 5 millions chess games with a computing grid.
During summer 2015, we processed 270 millions unique played positions using the Stockfish engine with a quite high depth (20). We populated a database of 1+ tera-octets of chess evaluations, representing an estimated time of 50 years of computation on a single machine.
Our effort is a first step towards the replication of research results, the supply of open data and procedures for exploring new directions, and the investigation of software engineering/scalability issues when computing billions of moves.
} 

\RRmotcle{jeu d'échecs, analyse de données, intelligence artificielle} 
\RRkeyword{chess game, data analysis, artificial intelligence} 
\RRprojets{DiverSE} 
 \RCRennes 

\begin{document}
\RTNo{479}
\makeRT 

\maketitle


\section{Introduction}

Millions of chess games have been recorded from the very beginning of chess history to the last tournaments of top chess players.
Meanwhile chess engines have continuously improved up to the point they cannot only beat world chess champions but also provide an authoritative assessment~\cite{CCRL,deepblue,stockfish,giraffe}.
The strengths of chess engines together with the availability of numerous chess games have attracted the attention of chess players, data scientists, and researchers during the last three decades. For instance professional players use chess engines on a daily basis to seek strong novelties; chess players in general confront the moves they played to the evaluation of a chess engine for determining if they do not miss an opportunity or blunder at some points.

From a scientific point of view, numerous aspects of the chess game have been considered, being for quantifying the complexity of a position, assessing the skills, ratings, or styles of (famous) chess players~\cite{van2005psychometric, HTCAWCC, HCPC,DBLP:conf/icaart/BiswasR15,DS,regan2015b}, or studying the chess engines themselves~\cite{haworth2007gentlemen,ISDCPS}.
Questions like "Who are the best chess players in history?" can potentially have a precise answer with the objective (and hopefully optimal) judgement of a chess engine.
So far numerous applications have been considered, such as methods for detecting cheaters~\cite{LEACDC,regan2012}, the computation of an intrinsic rating or the identification of key moments chess players blunder~\cite{ICR}.

A key issue for the research community is to gather a large dataset of chess games together with the judgement of chess engines~\cite{regan2015b}.
For doing so, scientists typically need to analyze millions of games, moves, and combinations with chess engines. Unfortunately it still requires lots of computations since (1) there are numerous games and moves to consider while (2) chess engines typically need seconds for fully exploring the space of combinations and thus providing a precise evaluation for a given position.
As a result and due to the limitation of computing storage or power, chess engines have been executed on a limited number of games or with specific parameters to reduce the amount of computation.

Our objective is to propose an open infrastructure for the large-scale analysis of chess games.
We hope to consider more players, games, moves, chess engines, parameters (e.g., the depth used by a chess engine), and methods for processing the overall data. With the gathering of a rich and large collection of chess engines' evaluations, we aim to (1) replicate state-of-the-art research results~\cite{regan2015b} (e.g., on cheat detection or intrinsic ratings); (2) provide open data and procedures for exploring new directions; (3) investigate software engineering/scalability issues when computing millions of moves;
(4) organize a community of potential contributors for fun and profit.

In this paper, we report our recent effort to analyse almost 5 millions chess games with a computing grid.
During summer 2015, we processed 270 millions of unique played positions using the Stockfish~\cite{stockfish} chess engine with a quite high depth (20).
Overall we populated a database of 1+ tera-octets of chess evaluations, representing an estimated time of 50 years of computation on a single machine.

Data analysts or scientists can use the dataset as well as the procedures to gather novel insights, revisit existing works, or address novel issues.
The lessons and numbers of our experience report can also be of interest for launching other large-scale analysis of chess games with other chess engines, games, and settings.






\section{Dataset and Chess Games}

 We gathered 4.78 million unique games publicly available on some Web repositories. In this section we report on some properties of the dataset.

 From a technical point of view, we parsed Portable Game Notation (PGN) files with a Java parser\footnote{https://github.com/jvarsoke/ictk} and pgn-extract\footnote{https://www.cs.kent.ac.uk/people/staff/djb/pgn-extract/}.
 As a preprocess we notably eliminated a significant number of duplicated games with pgn-extract.
 We used the \textsf{R} programming language to produce plots and results.

Interestingly we can confront our results to a previous attempt by Randal Olson that explored a dataset of over 650,000 chess tournament games (see a series of blog posts\footnote{ http://www.randalolson.com/2014/05/24/chess-tournament-matches-and-elo-ratings/}).
 We consider similar properties (such as distribution of Elo ratings) as well as additional ones (such as most represented events).
 In our case we have much more games and it is worth comparing our numbers to Olson's results.
  Our objective was to have a better understanding of the quality of the dataset,
  since some factors can have a negative impact on their exploitations by chess engines.


\subsection{Elo Ratings}

\begin{figure*}[h!]
     \centering
     \subfloat[Distribution of Elo ratings]{\label{fig:ratings}\includegraphics[width=0.4\textwidth]{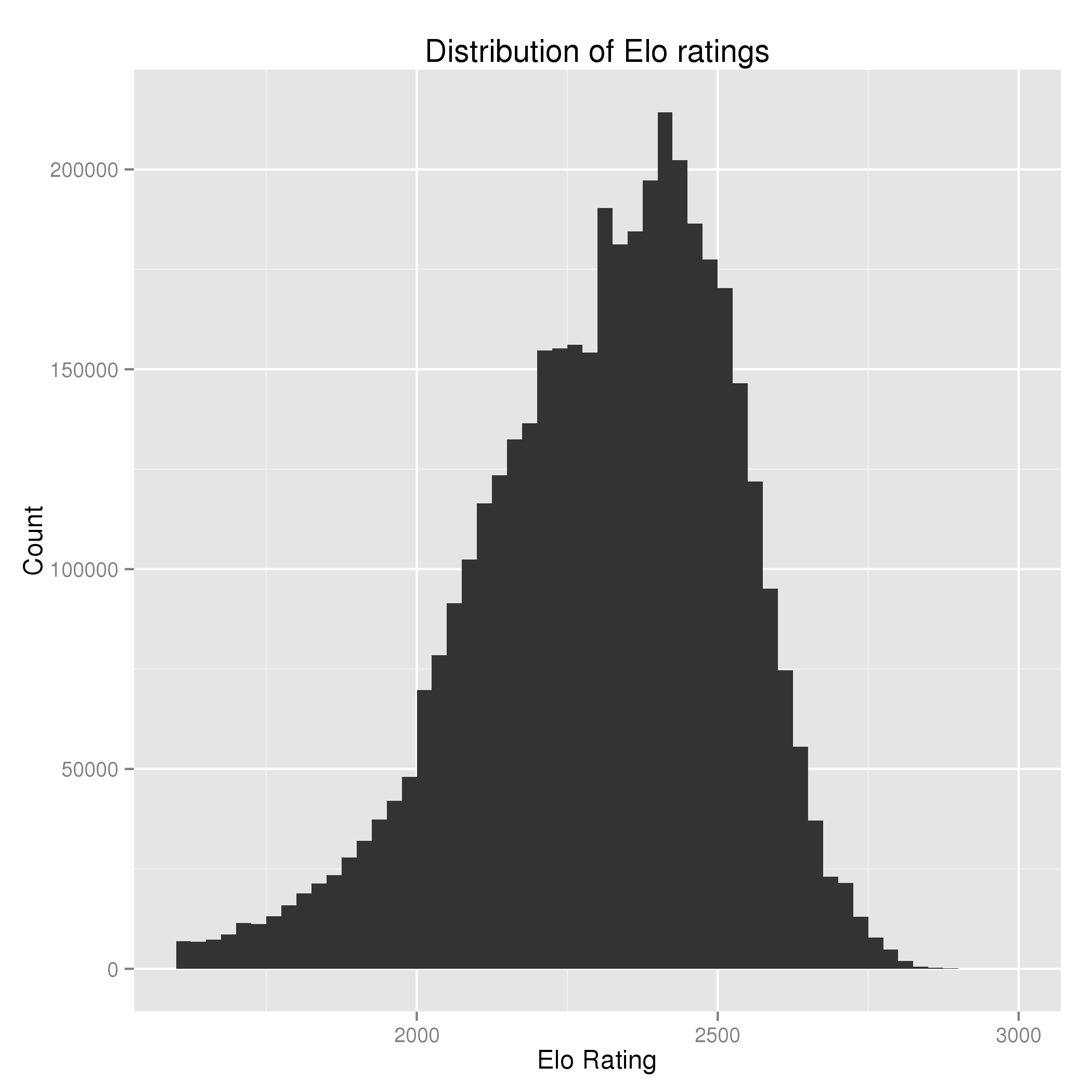}}
   \hspace{3pt}
     \subfloat[Differences of Elo]{\label{fig:diffratings}\includegraphics[width=0.4\textwidth]{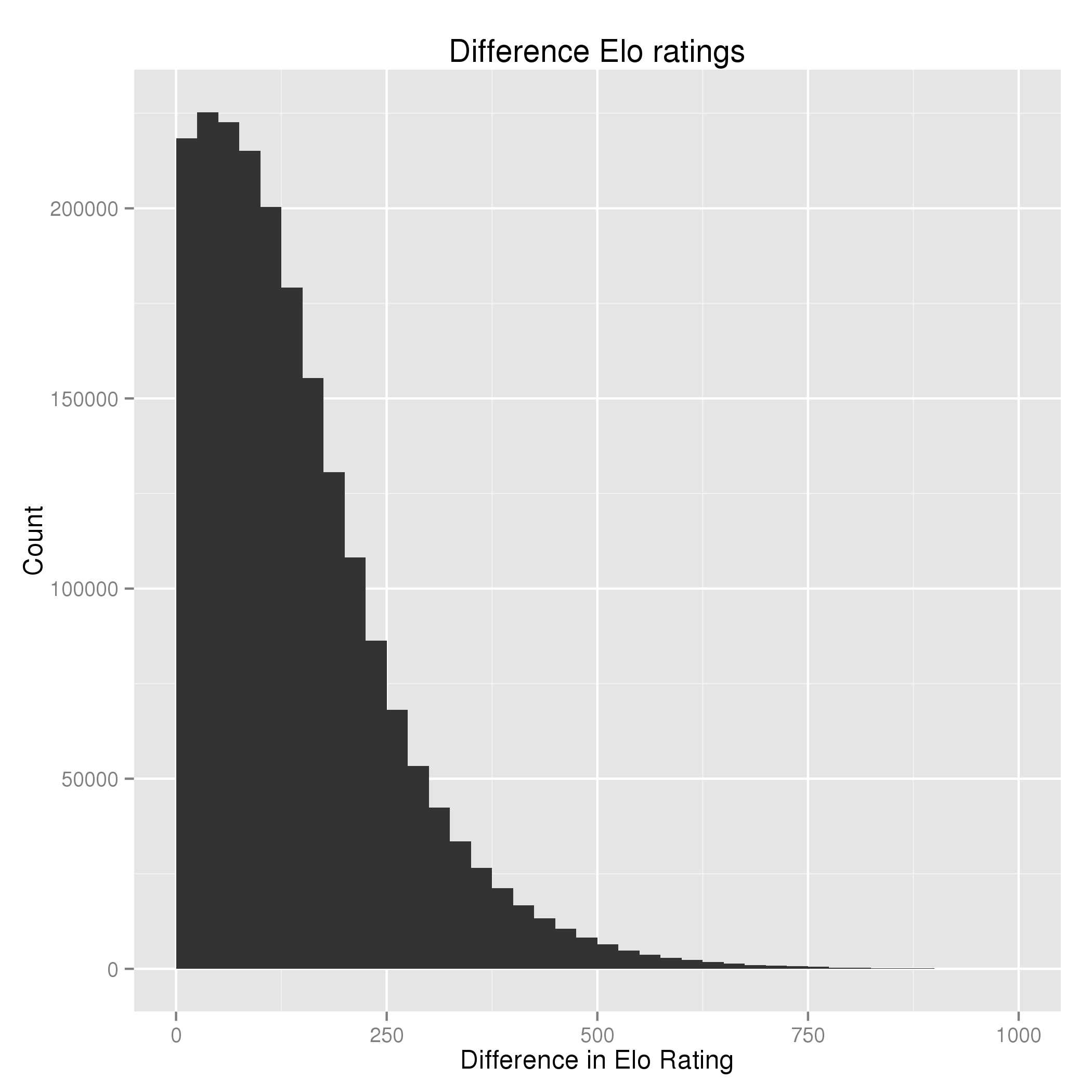}}
     \caption{Distribution and differences of Elo\label{fig:elorates}}
 \end{figure*}

The Elo rating system is intensively used in chess for calculating the relative skill levels of players.
Figure~\ref{fig:ratings} gives the distribution of Elo ratings thanks to PGN headers.
It should be noted that the Elo rating was adopted by the World Chess Federation (FIDE) in 1970. As such oldest games do not appear in Figure~\ref{fig:elorates}.

Players with \textgreater\,2000 Elo rating are considered as experts in Chess; between 2200 and 2400, players are national masters; above 2400, international masters and above 2500 international grand masters.
Figure~\ref{fig:ratings} shows that most records come from games with players \textgreater\,2200 Elo, with a large proportion played by international masters, grand masters, and top players.
 %
 In general, the difference of Elos between two opponents is quite close (\textless\,100), see Figure~\ref{fig:diffratings}. It is in line with what observed in chess tournaments (e.g., round-robin or swiss systems).



\subsection{Ply per Games}

\begin{figure}[h!]
    \centering
   \includegraphics[width=0.4\textwidth,]{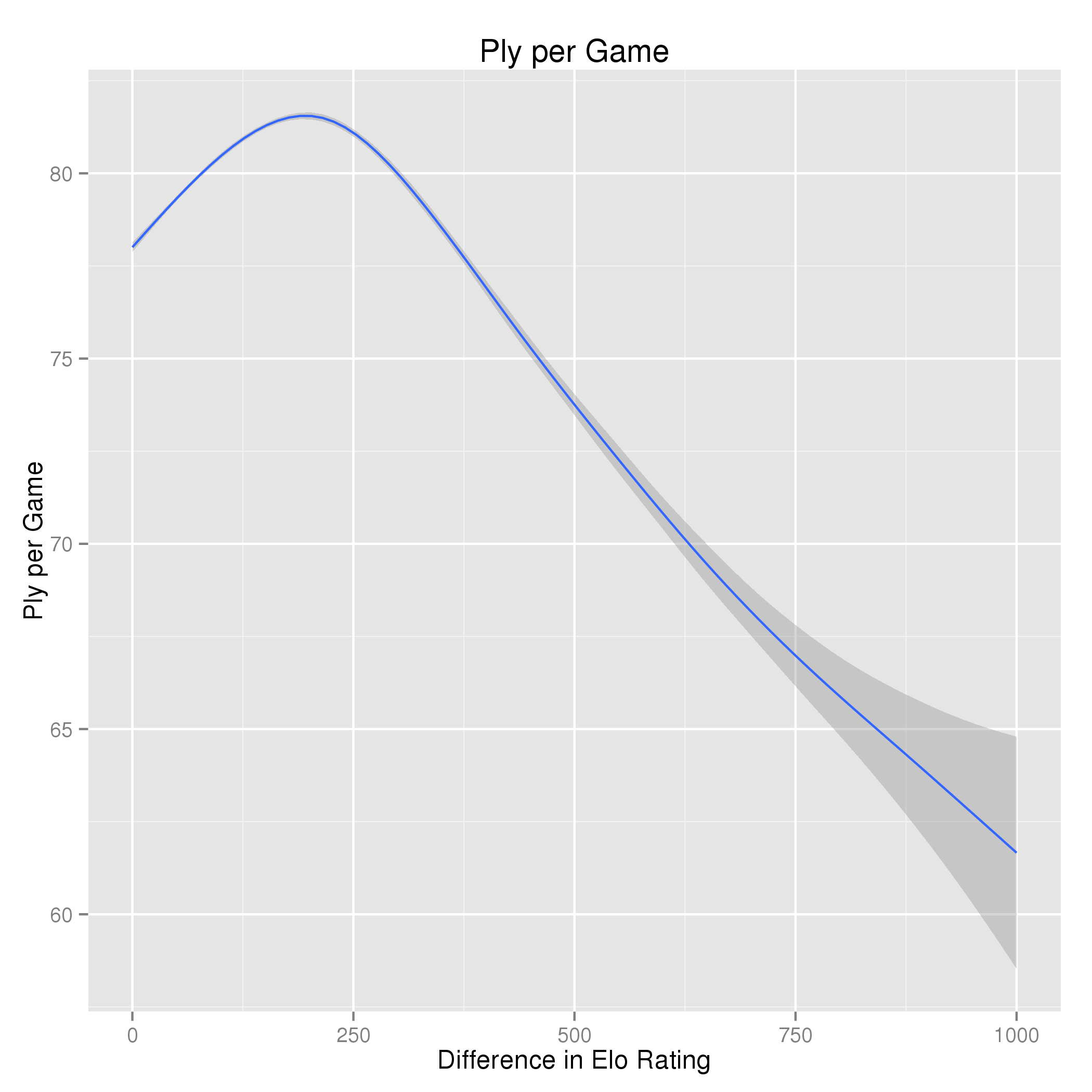}
    \caption{Ply per Game w.r.t. difference in Elo rating}
    \label{fig:plyelo}
\end{figure}

Our 4.78 million unique games have a mean of 80 ply per game (40 moves).
Figure~\ref{fig:plyelo} shows that the number of moves follows two trends:
With \textless 200 Elo difference, the number of ply\footnote{A ply refers to one turn taken by one of the chess players} per game varies between 78 and 85.
It is not clear why there is a slight increase.
We can formulate the assumption that some games end early with a draw in case the Elo difference is closed.
With a larger Elo difference, the number of Ply starts decreasing until 60 ply per game. It is quite intuitive since strongest players gain the upper hand quickly over their opponents.

The number of ply in some games can be suspicious or presents limited interests:
\begin{itemize}
\item Klip,H  (2305) - Bottema,T (2205) \textsf{1. e4 f6 2. d4 g5 3. Qh5\# 1-0} (1990)
\item Landa,K (2678) - Grall,G (1812) \textsf{1. e4 e5 2. Bc4 Bc5 3. Qh5 Nf6 4. Qxf7\# 1-0} (2007)
\item Strekelj,V (1843) - Kristovic,M (2328) \textsf{1. f3 e5 2. Kf2 d5 3. Kg3 Bc5 4. Nc3 Qg5\# 0-1} (2011)
\end{itemize}


On the other hand, we have very long games, for example:
\begin{itemize}
\item Sapin - Hyxiom (2003-01-01) 600 plies
\item Felber,J (2150) - Lapshun,Y (2355) (1998-09-05) 475 plies
\item Sanal,V (2286) - Can,E (2476) (2012-03-29) 456 plies
\end{itemize}


\subsection{Elo Difference and Winning Chance}

\begin{figure*}
\centering
\includegraphics[width=0.5\textwidth]{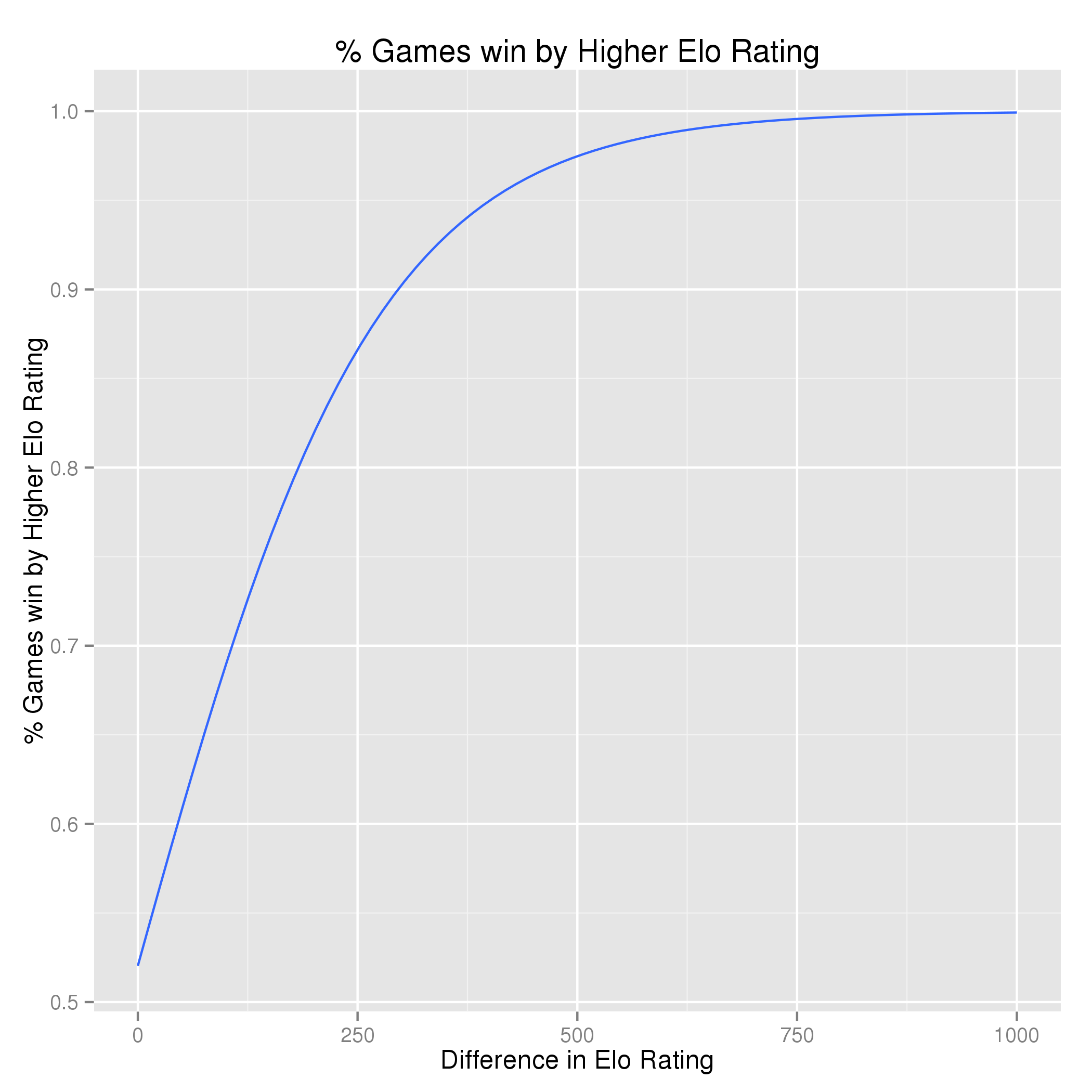}
\caption{Elo difference and games win (similar probabilities graph/outcomes of Elo rating system)\label{fig:elowin}}
\end{figure*}

The difference in Elo rating strongly predicts the winner of the game. With a difference of 100, the chance for the higher player to win is 70\% -- not counting draws. The graph of Figure~\ref{fig:elowin}
is in line with the probabilities and expected outcomes of the Elo rating system.
In general our dataset provides PGN headers information (Elo, winner) we can trustfully exploit in the future. We have also some exceptional records/anecdots:
\begin{itemize}
\item \textbf{Vera Gonzalez,J (1551)} --- Hernandez Carmenates,Hold (2573):    1-0 (Elo difference: 1022)
\item \textbf{Freise,E (2018)} --- Anand,V (2794): 1-0 (Elo difference: 776)
\item \textbf{Sikora,J (1833)} --- Movsesian,S (2710): 1-0  (Elo difference: 877)
\end{itemize}

\begin{figure*}[h!]
    \centering
    \subfloat[White players, Elo rating, and games win]{\label{fig:whitecolor}\includegraphics[width=0.45\textwidth]{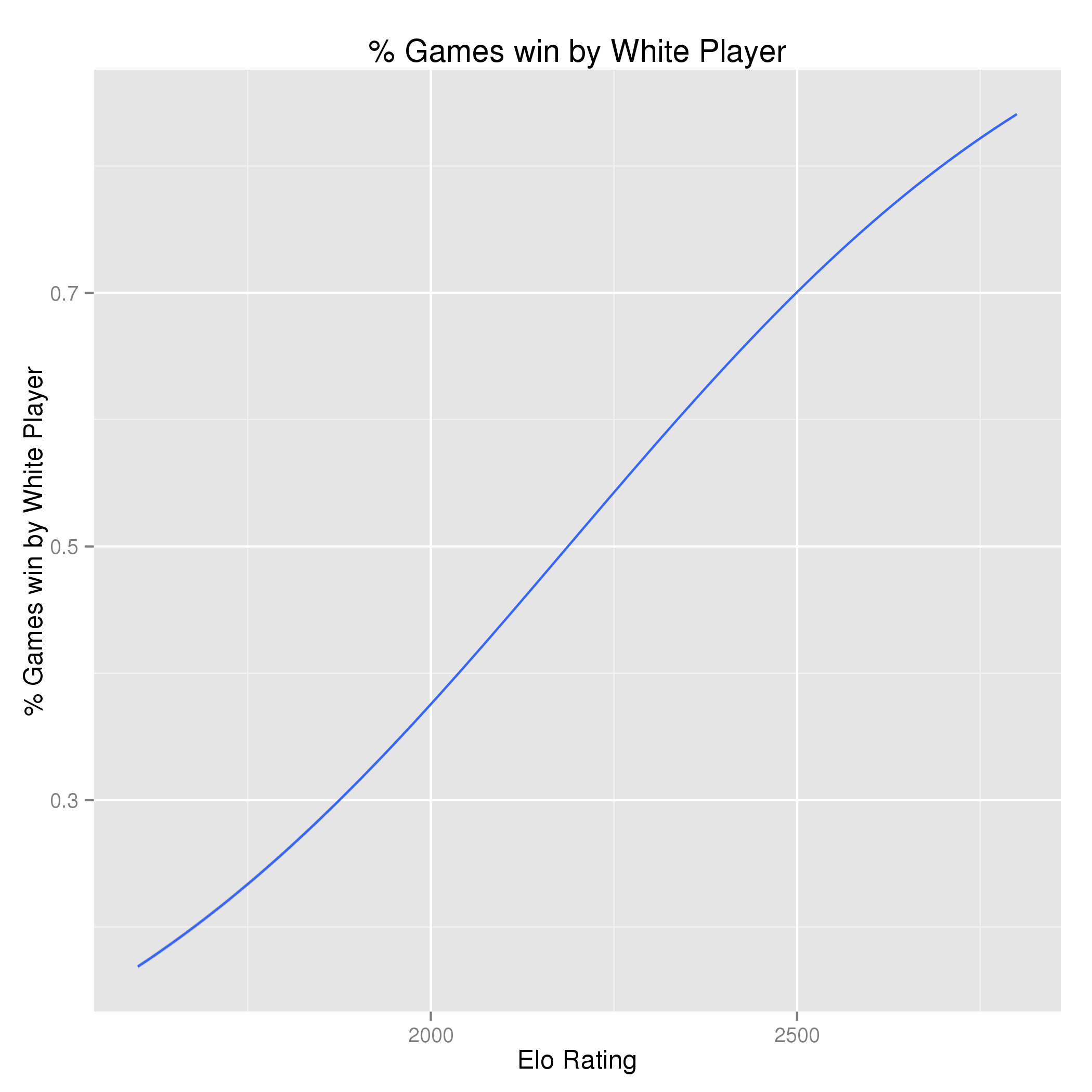}}
	\hspace{3pt}
    \subfloat[Games win, color, and Elo]{\label{fig:whitecolor2}\includegraphics[width=0.5\textwidth]{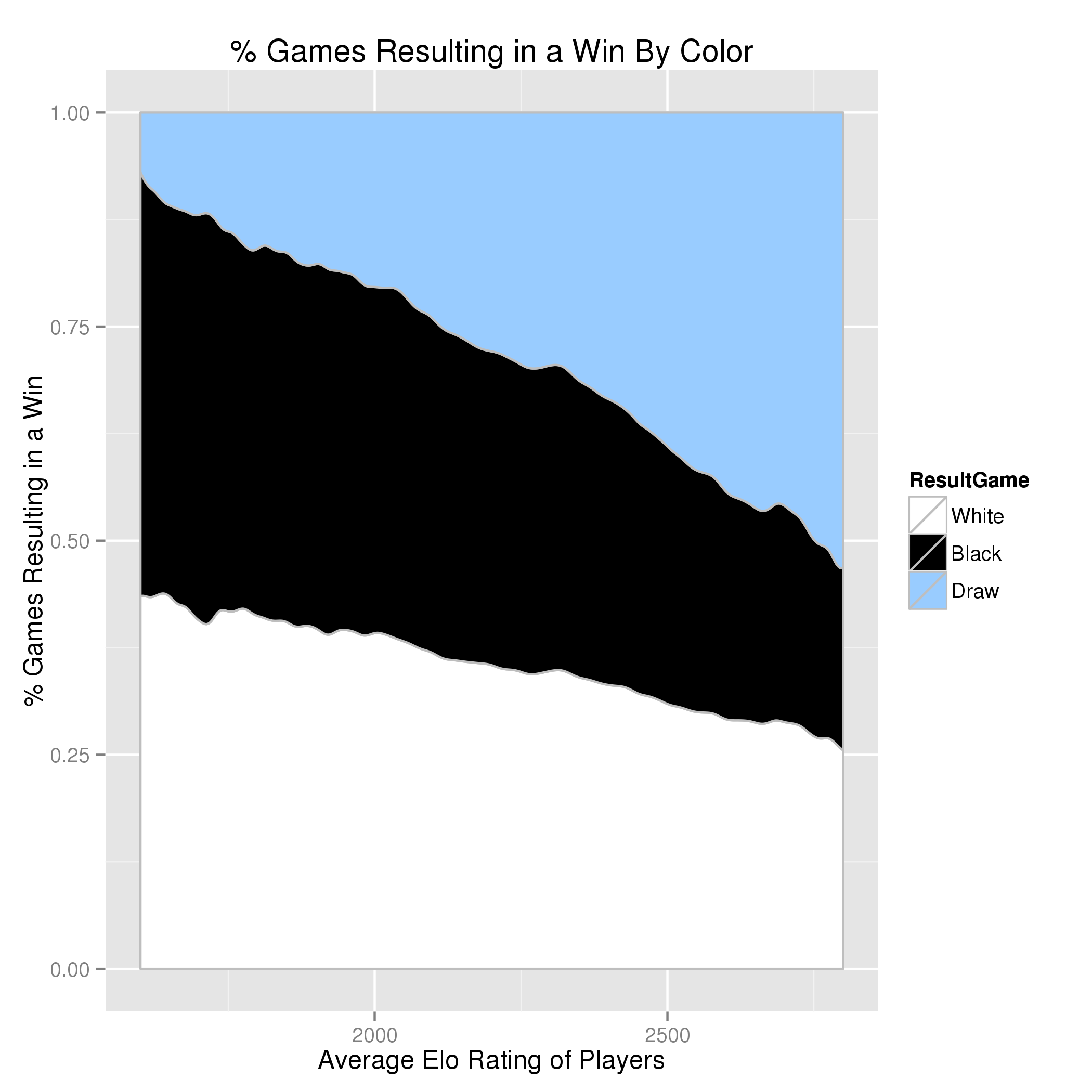}}
    \caption{Colors and percentage win}
\end{figure*}

\subsection{Colors and Percentage Win}

Figure~\ref{fig:whitecolor} shows that the percentage games won by white players depends of Elo Rating.
Below 2200 Elo, chances to lose are higher (even with white colors) because such players face to better opponents and simply have less chance to win.
On the other side, the percentage to win for a 2500 white player is 70\% -- not counting draws.
 Specifically, we have 120K games with white player having a Elo rating \textgreater 2600: They win more than 55K games, draw 50K and lose only 15K games.
Figure~\ref{fig:whitecolor2} shows the importance of having white pieces.
 For low Elo ratings, the draw rarely happens whereas it happens half the time for stronger players.
All these results are coherent with what is usually observed in chess.

\subsection{First moves and Openings}

\begin{figure*}[h!]
    \centering
    \subfloat[]{\label{fig:firstmoves1}\includegraphics[width=0.45\textwidth]{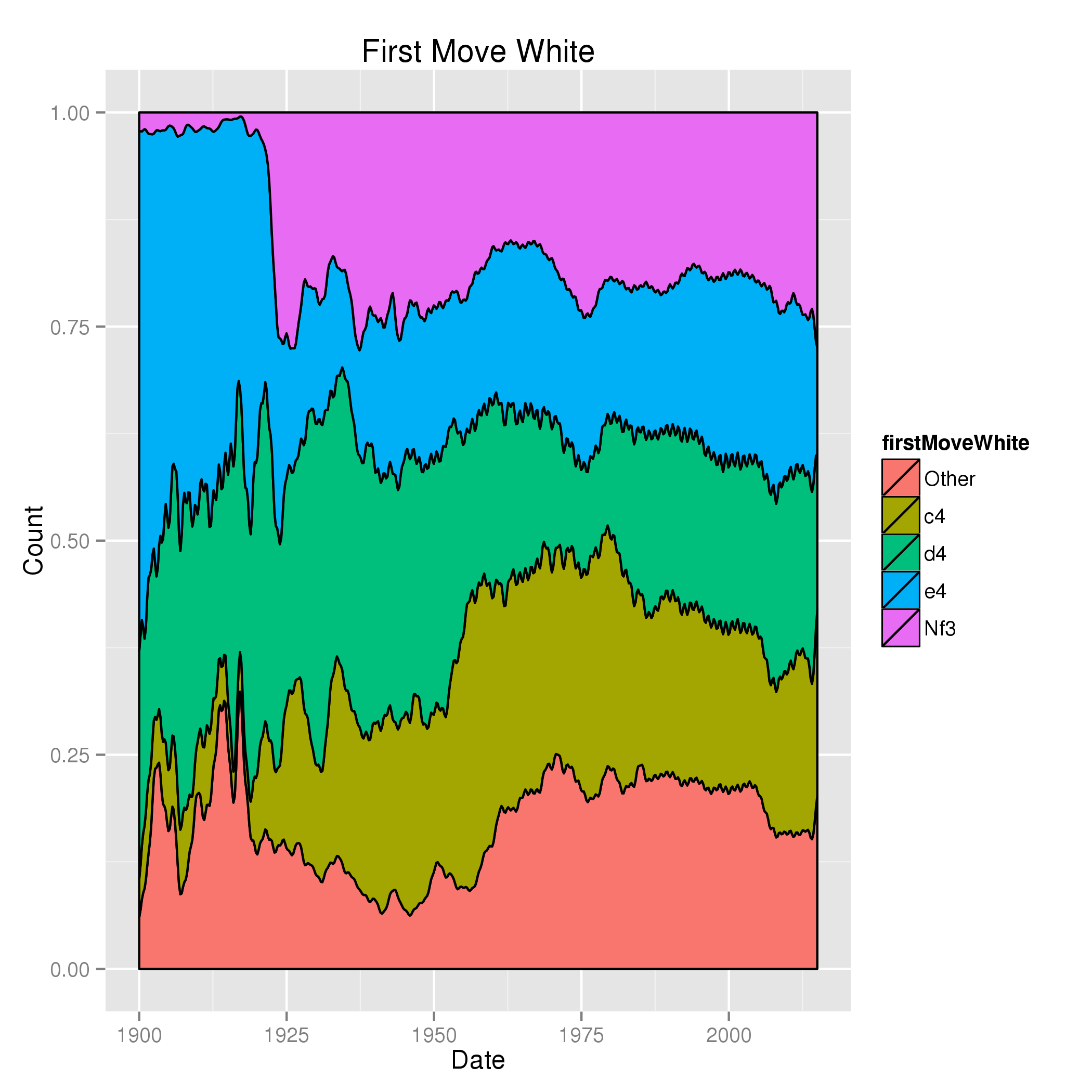}}
	\hspace{3pt}
    \subfloat[]{\label{fig:firstmoves2}\includegraphics[width=0.45\textwidth]{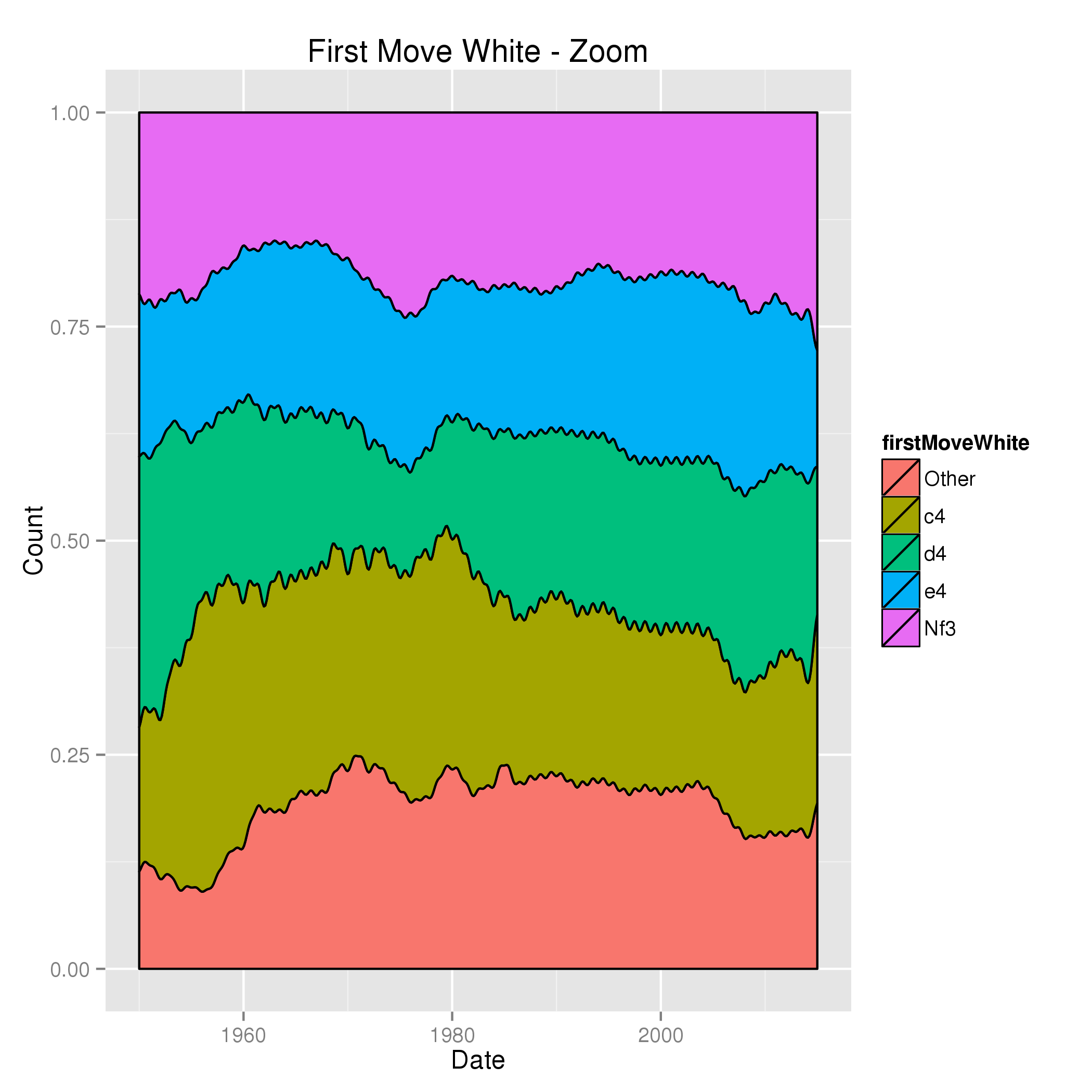}}
    \caption{First moves (white) depending on the dates}
\end{figure*}

Figure~\ref{fig:firstmoves1} shows that the period 1900 -- 1960 was more prone to fluctuations/experiments for the choice of the first moves since the openings theory was not developed.
d4, e4, Nf3, and c4 are the most popular first moves to start a game (see Figure~\ref{fig:firstmoves2}).
It should be noted that for the period 1960 -- 2015 the top first moves slightly change over the time and tend to stabilize.
It is consistent with existing databases and current practices.


\subsection{Distribution of Date Match}

\begin{figure*}[h!]
    \centering
    \subfloat[Years of games]{\includegraphics[width=0.45\textwidth]{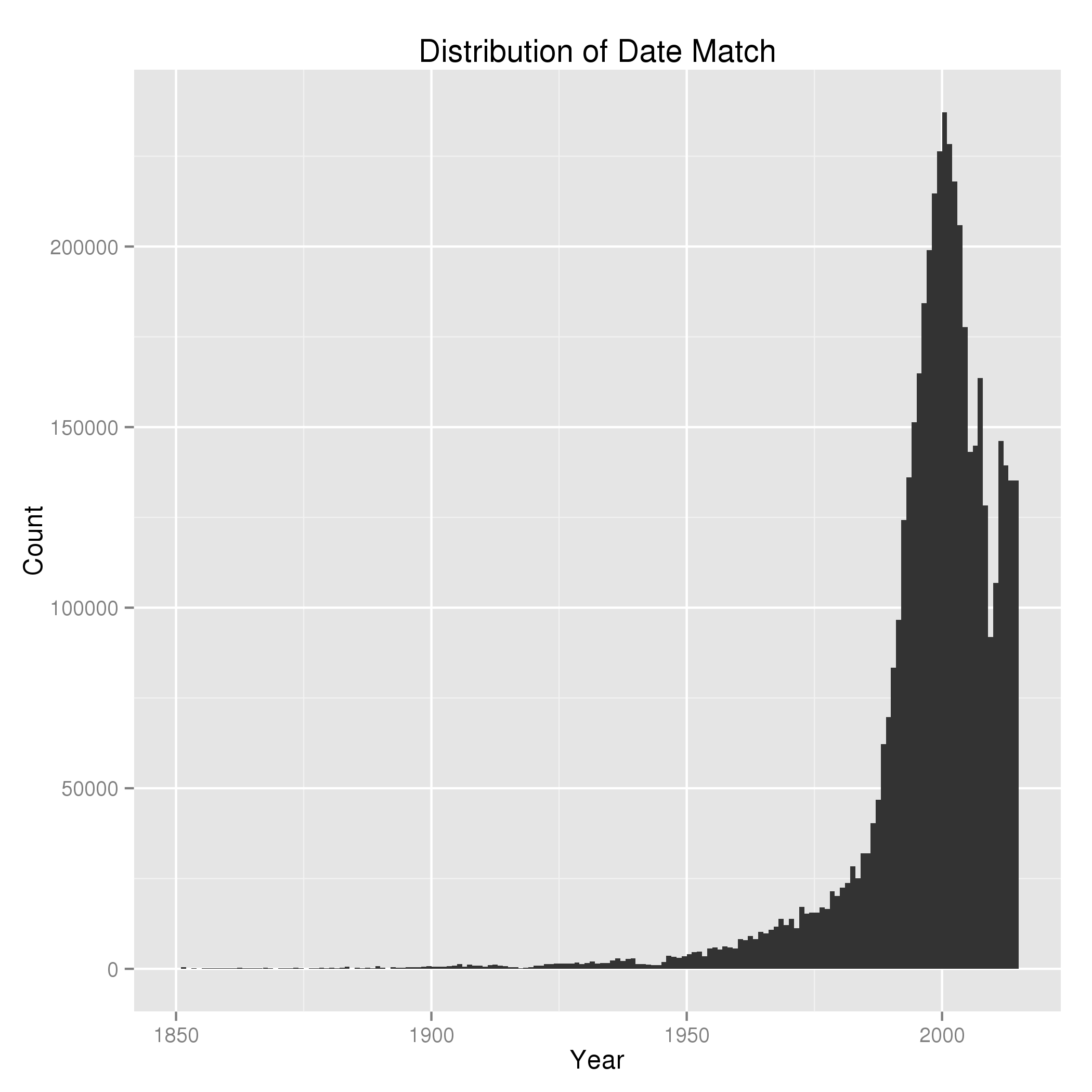}}
	\hspace{3pt}
    \subfloat[Ply per game depending on the date]{\includegraphics[width=0.45\textwidth]{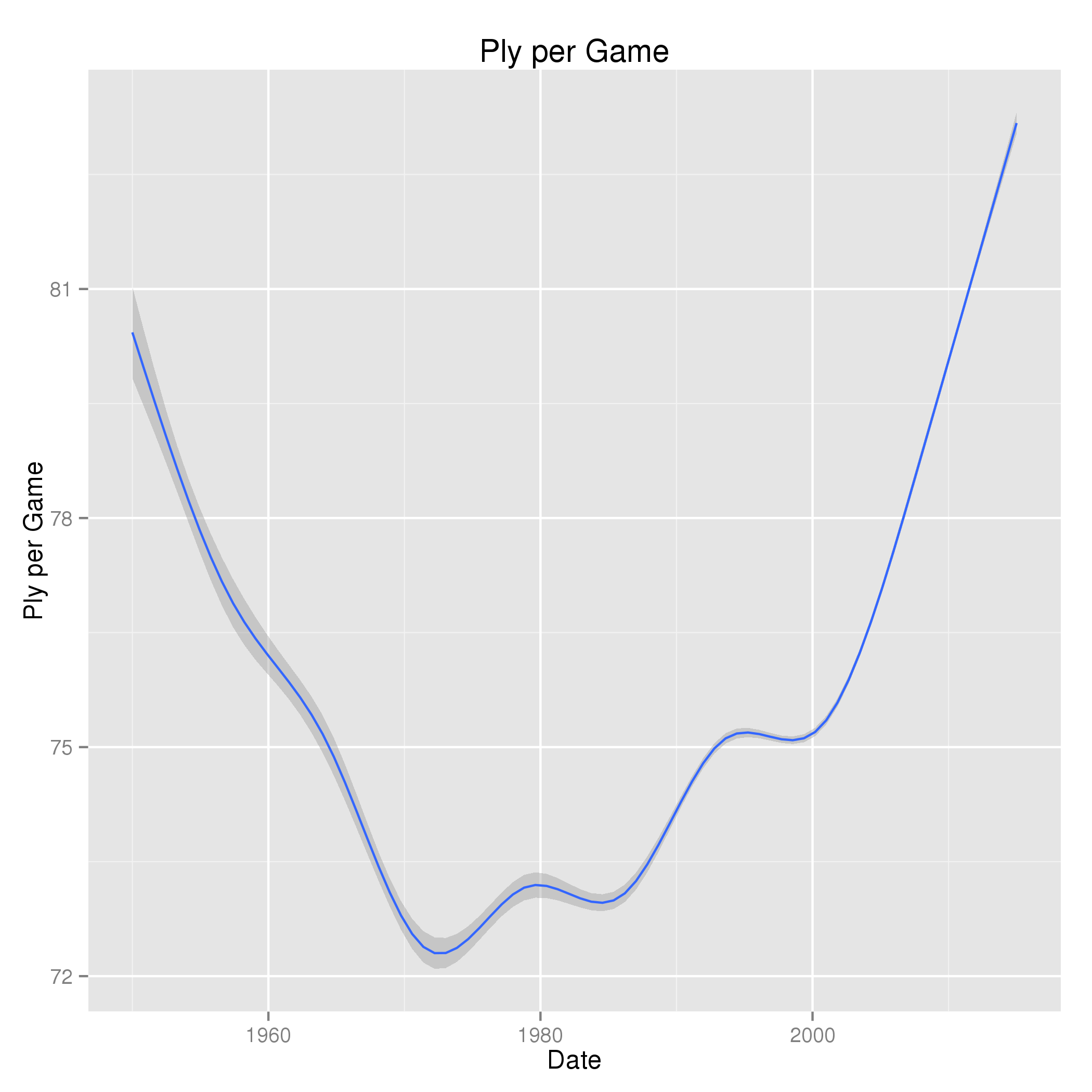}}
    \caption{\label{fig:years}Games and years/dates}
\end{figure*}

Before 1950s, only interesting games are recorded and saved for posterity.
Now, amateurs and professionals can easily share PNG files (see, e.g., TWIC). The number of records explodes in the 2000s. A noticeable curve is observed in Figure~\ref{fig:years}: 81 plys in 1950s, 72 plys in 1970 and about 84 today. We have 679K games recorded after 2010 and the average is 82 ply per game.

\begin{figure*}[h!]
    \centering
    \subfloat[Percentages games win by white player depending on the date]{\label{fig:yearswhite}\includegraphics[width=0.45\textwidth]{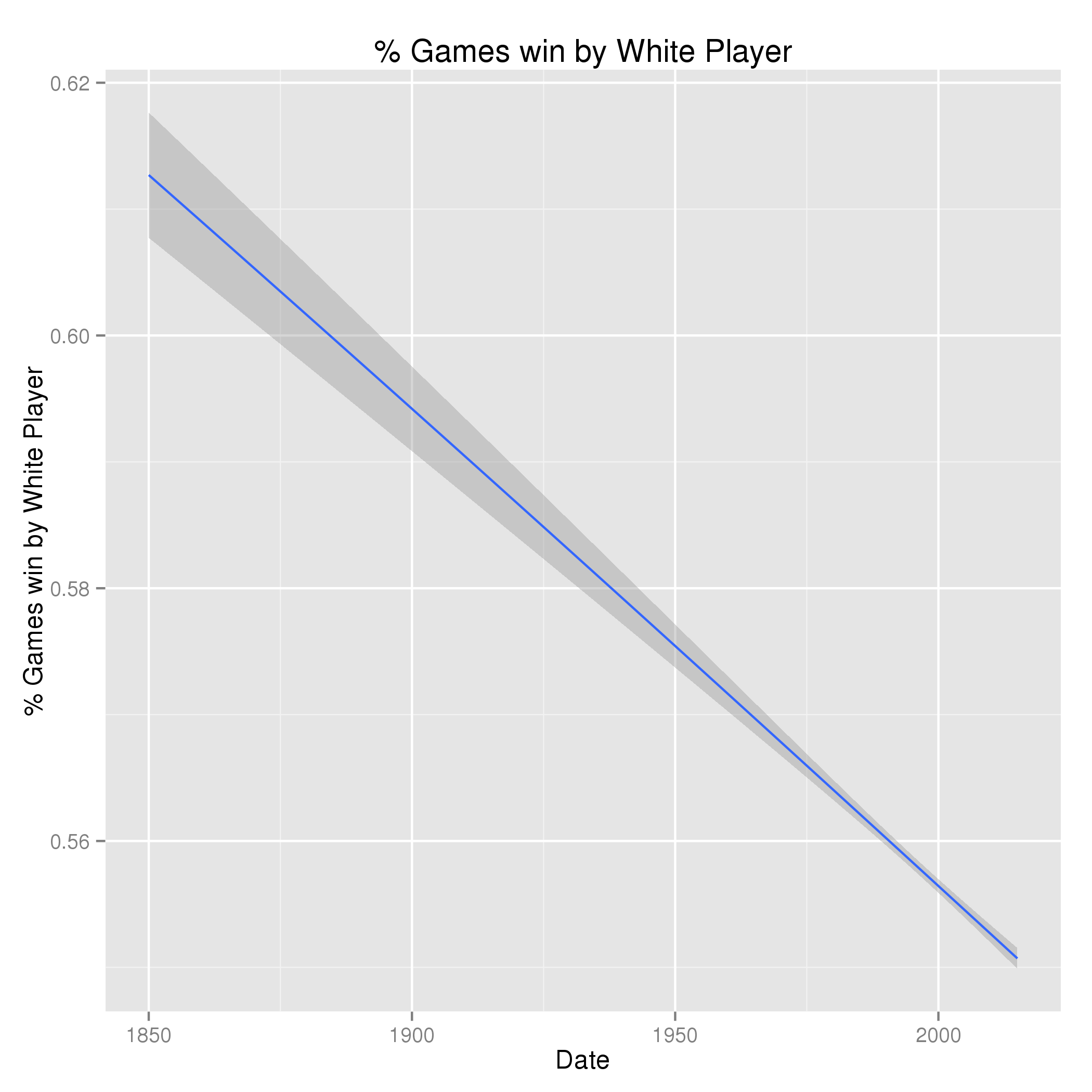}}
	\hspace{3pt}
    \subfloat[Percentages games win by color depending on the date]{\label{fig:colorstime}\includegraphics[width=0.5\textwidth]{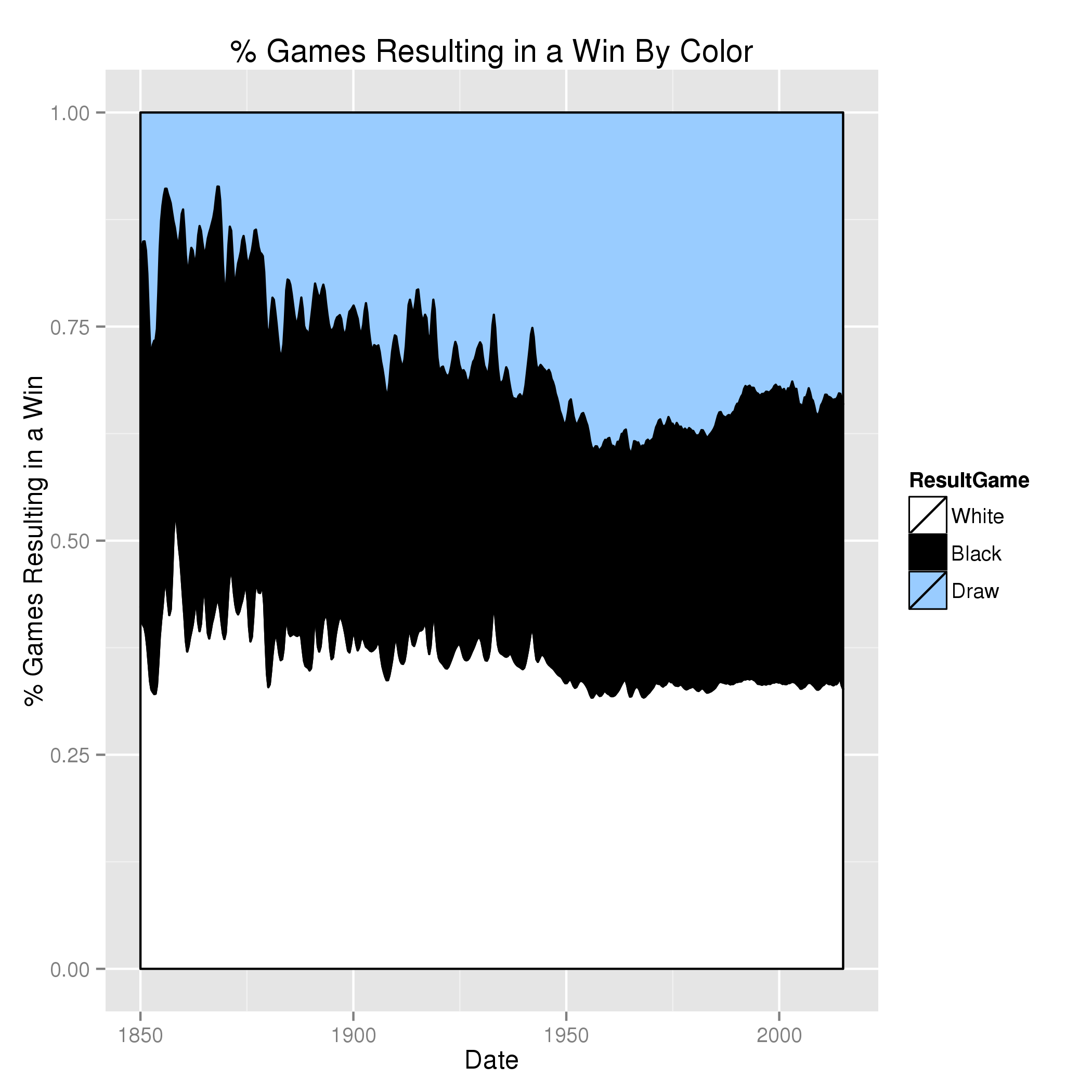}}
    \caption{Percentages games win by color w.r.t. dates}
\end{figure*}

Figure~\ref{fig:yearswhite} and Figure~\ref{fig:colorstime} show that the percentage win by white players tends to decrease steadily even if the first-move still provides a small advantage.

%




\subsection{Comparison with Olson's dataset}

Compared to Olson's results, we observe similar properties like:
\begin{itemize}
\item the distribution of Elo ratings and difference of Elo ratings between two opponents
\item the importance of white and first-move advantage
\item the increase of draws when chess experts are involved
\item the proportion of first moves and openings
\item the fact Elo ratings tend to predict game outcome
\end{itemize}

We also observe some differences; we comment two of them here. First, the number of moves depending on difference Elo Rating (see Figure~\ref{fig:plyelo}, page~\pageref{fig:plyelo}): Olson's dataset does not exhibit an increase curve between 0 and 200 difference Elo rating. Moreover the number of ply is slightly higher all along the graph.
 Second, the proportion of games win by white players depending on the date (see Figure~\ref{fig:yearswhite}, page~\pageref{fig:yearswhite}). We concur with the conclusion that having white pieces is an advantage for all periods. However we also observe a linear decrease that is not apparent in Olson's dataset, especially for old periods.

 Our conclusion is that there is no fundamental difference, i.e., the number of games considered in the two datasets can explain such differences.

\subsection{Summary}

Appendix~\ref{app:misc} provides further results related to kinds of moves (promoted rates,
queen side castling rates, etc.).
Several other information can be extracted from the dataset as well.
 Our results so far suggest that (1) the database contains numerous interesting games with rather strong players; (2) headers information such as Elo rating or results of the games are coherent.
Though some games can certainly be removed or corrected,
we consider the dataset is representative of existing chess databases and consistent with chess practices and trends. We also obtain similar properties than in other datasets or databases.
Finally, the number of games (almost 5 millions) is significant.

\section{Large-scale Analysis of Chess Games with Chess Engines}

We have now a better understanding of our database of chess games.
We consider that the properties of the dataset are reassuring and justify the analysis of all games by chess engines -- our original motivation.
In a sense our next objective is to produce a \emph{dynamic} analysis of chess games (by opposition to a static analysis as we made in previous section).
The exploitations of the chess engines analysis (for blunder detection, players' ratings, etc.) are left as future work.
In this section and preliminary report, we describe how we performed a large-scale analysis.

\subsection{Analysis process}

\textbf{270 millions FEN positions.} We encoded each game position using the Forsyth-Edwards Notation (FEN) notation. 
Some positions are equals (including the context leading to the position): a FEN encoding allows us to detect such equality. The underlying idea is that a chess engine is executed only one time per equal position.
We also observe that some positions are theoretical openings and quite well-known, presenting limited interests for an analysis: we used Encyclopaedia of Chess Openings (ECO) code to detect such positions.
Our dataset originally exhibits 380 millions positions.
By exploiting FEN encoding and ECO classification, we only had to consider 270 millions positions.
With these simple heuristics, we drastically reduced the number of games to analyze with chess engines.

\textbf{Stockfish chess engine depth=20, multipv=1.} We used Stockfish~\cite{stockfish} (version 6) to analyze all FEN positions.
Stockfish is an open source project and a very strong chess engine -- one of the best at the time of writing~\cite{CCRL}.
Other researchers have also considered Stockfish in prior works.
We used the UCI protocol to collect analysis.
 Importantly we set the depth to 20. It is a quite high depth (prior works usually set lower depths because of the computation cost). We also set multipv\footnote{UCI chess engines like Stockfish support multi best line or k-best mode. It means they return the k best moves/lines. When k=1, Stockfish returns the best line and evaluation. The increase of multipv is possible and has practical interests (see, e.g.,~\cite{regan2015b}), but it has also a computational and storage cost.} to 1.

\textbf{Structuring data.} We used a simple database schema (see Figure~\ref{fig:olsonelo}, page~\pageref{fig:olsonelo}).
PGN headers informations are stored and structured for retrieving games, positions, players, etc.
For each position (FEN), we associate the score and log (multipv=1) computed by Stockfish.
We developed several proof-of-concepts to validate the schema: \url{https://github.com/ChessAnalysis/chess-analysis-database}. For instance, we can gather all positions of a game and depict the scores' evolution.

\textbf{Distributing the computation.} Our experiments suggested that it takes about 6 seconds to analyze a FEN on a basic machine.
The use of a single machine was simply not an option since we have to analyze 270 millions position.
It would require $270 * 10^{6} * 6 = 1620 * 10^{6}$ seconds, 450K hours, 18K days, and around 50 years of computation. The third step of our process was thus to distribute the computation on a cluster of machines.
We used IGRIDA\footnote{\url{http://igrida.gforge.inria.fr/}}, a computing grid available to research teams at IRISA / INRIA, in Rennes.
The computing infrastructure has 125 computing nodes and 1500 cores.

\textbf{Computational and storage cost.} We split the FEN positions for distributing the computation on different nodes. We processed in batch (without user intervention) and the analysis was incremental.
We used in average 200+ cores during night and day during 2 months.
We gathered around 1,5 tera-octets of data (FEN logs).

\subsection{Conclusion and Future Work}

Our experience showed that it is practically feasible to analyze around 5 millions chess games with state-of-the-art chess engines like Stockfish.

Our next step is naturally to process and exploit data for either replicating existing works or for investigating new directions~\cite{regan2015b}. The amount of analysis we have collected is superior to what have been considered so far in existing works: We expect to gain further confidence in existing results or methods (e.g., for players' ratings, blunders detection, influence of depth engines, etc.). Another direction is to collect more data. We used Stockfish with depth=20 and multipv=1 (single-pv mode).
It has some limitations; for example, some methods/applications require multipv mode.
We can rely on other chess engines. We can also use different settings for Stockfish (e.g., a higher depth and multipv). A possible threat is that the computation and storage cost can be very important.

We believe more data can help to better assess methods based on chess engines.
From this perspective, we are happy to share our results with scientists, chess experts or simply data hobbyists. We hope to confront methods or interpretations and have different perspectives on data.
More information can be found online: \url{https://github.com/ChessAnalysis/chess-analysis}

Our long-term goal is to better understand the underlying beauty and complexity of chess thanks to the incredible skills of contemporary chess engines.
The rise of computing power, software, and chess data can also be seen as an opportunity to address very old
questions in the fields of cognitive and computer science (e.g., artificial intelligence, computational complexity, data and software engineering).

\textbf{Acknowledgement.} We would like to thank our colleagues at DiverSE (\url{http://diverse.irisa.fr}) for their discussions and feedbacks.




\bibliographystyle{plain}
\bibliography{chess}


\newpage
\appendix

\section{Database schema}
\label{sec:database}

 \begin{figure*}[h!]
\centering
    \includegraphics[width=1.0\textwidth]{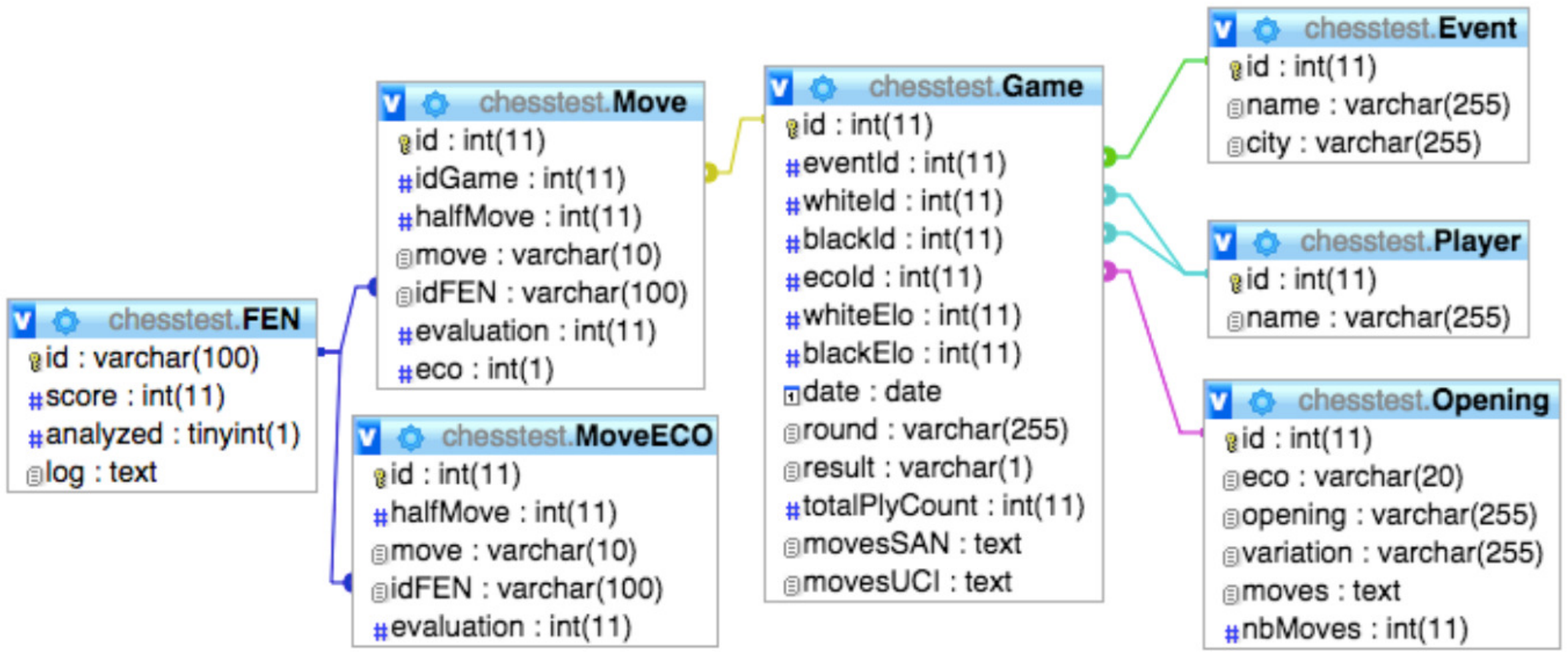} 
    \label{fig:olsonelo}
\end{figure*}

\newpage
\section{Misc: checkmates, captured pieces, promoted rates, and kinds of moves}
\label{app:misc}

\begin{figure}[h!]
\centering
   \includegraphics[width=0.6\textwidth]{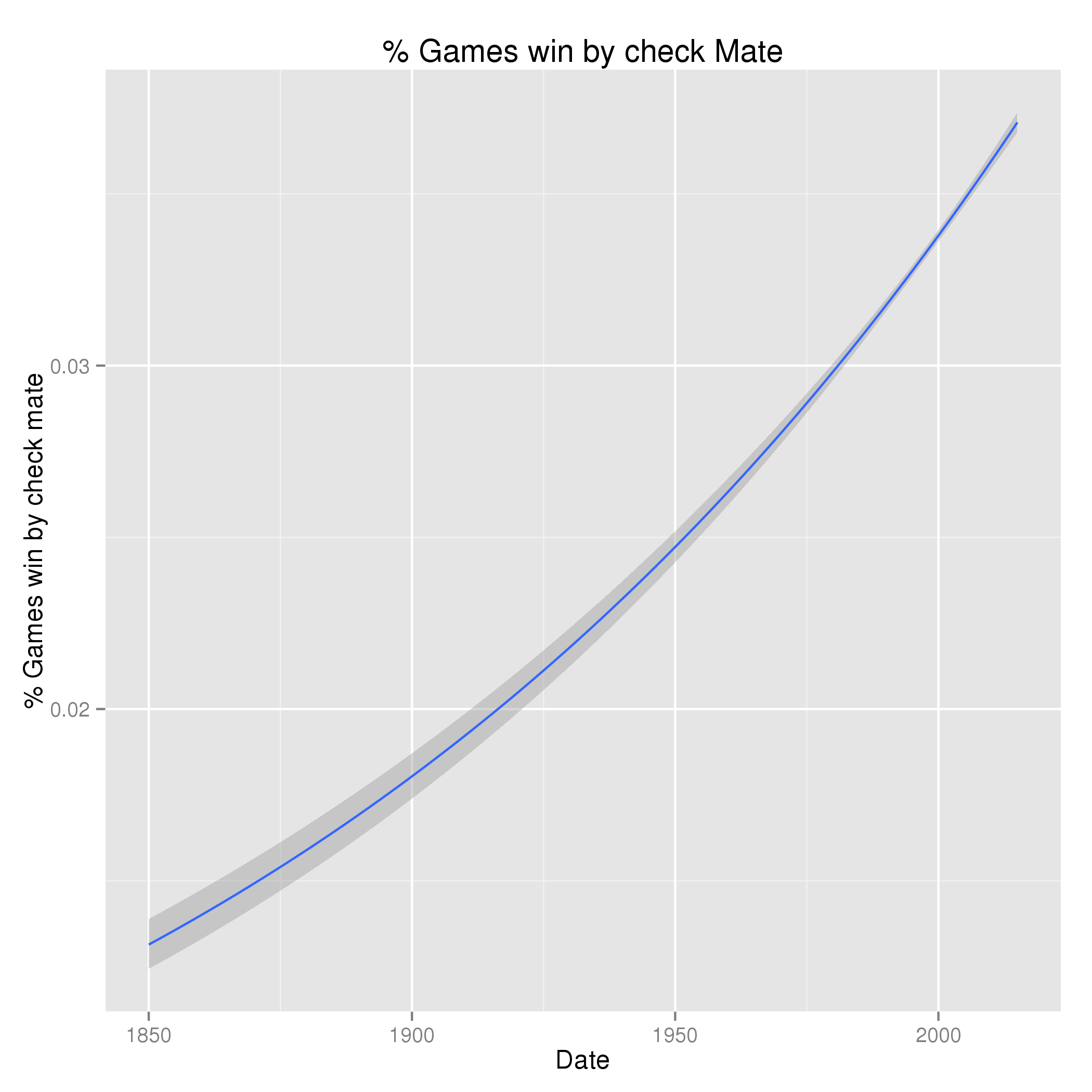}
 \caption{
 There are only a few checkmates during games because players typically resign when they realize the defeat is near.
 The slight increase is surprising and deserves more investigations (e.g., a possible explanation is the inclusion of recent rapid games like Blitz in the dataset).
 \label{fig:checkmates}}
\end{figure}

\begin{figure}
\centering
  \includegraphics[width=0.6\textwidth]{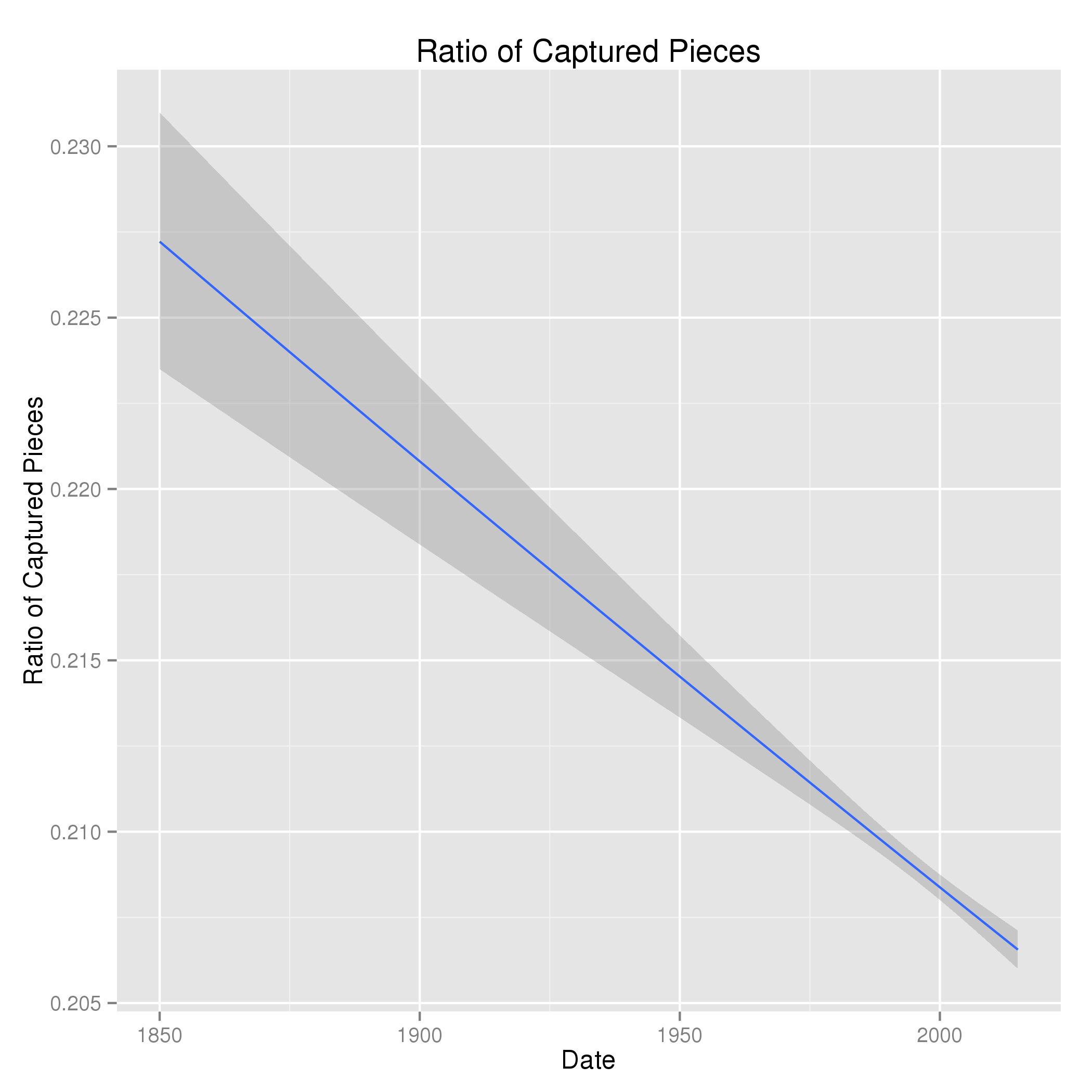}
  \label{fig:cpatures}
  \caption{The ratio of captured pieces during a game has slowly decreased}
\end{figure}

\begin{figure}
\centering
 \subfloat[]{\includegraphics[width=0.35\textwidth]{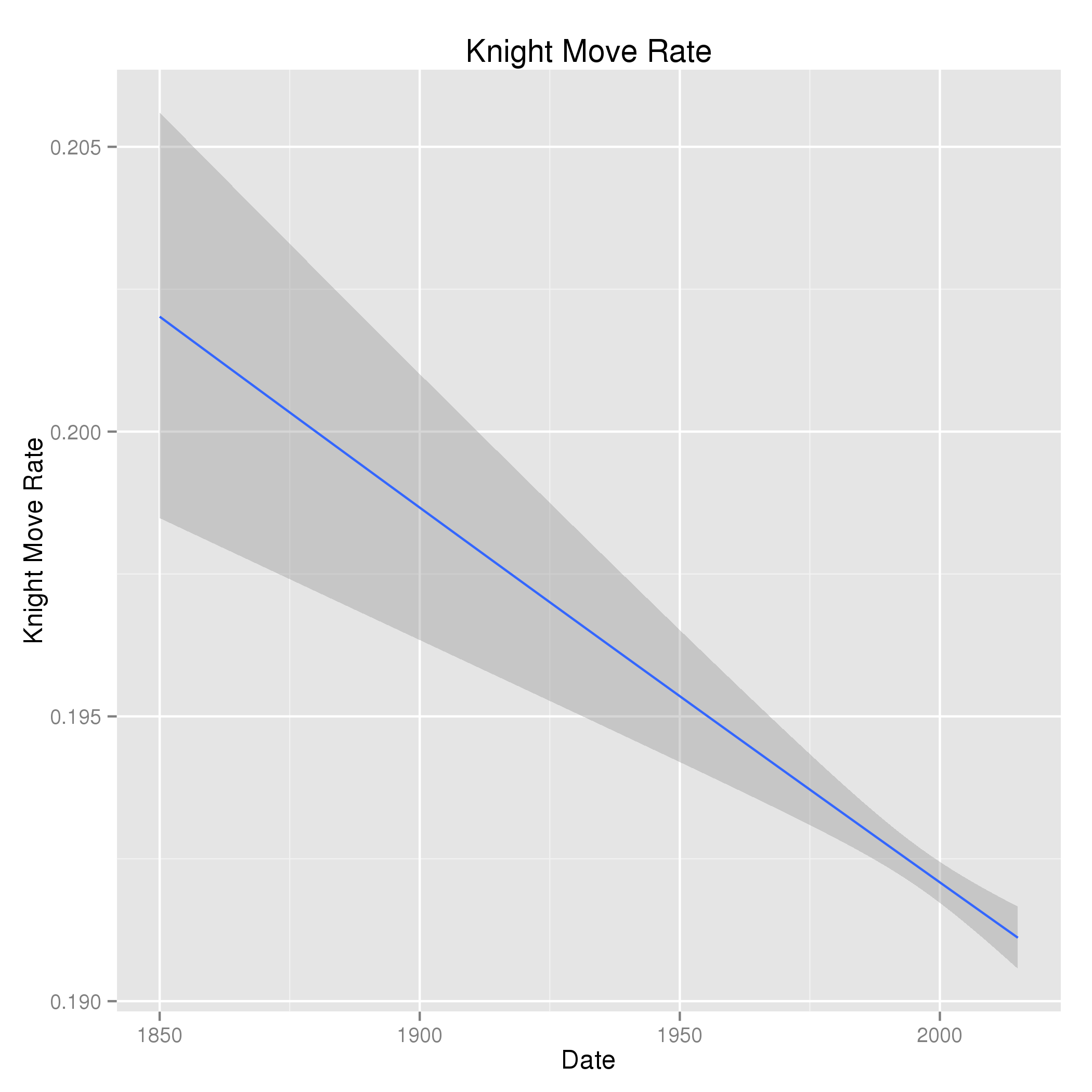}}
 \subfloat[]{\includegraphics[width=0.35\textwidth]{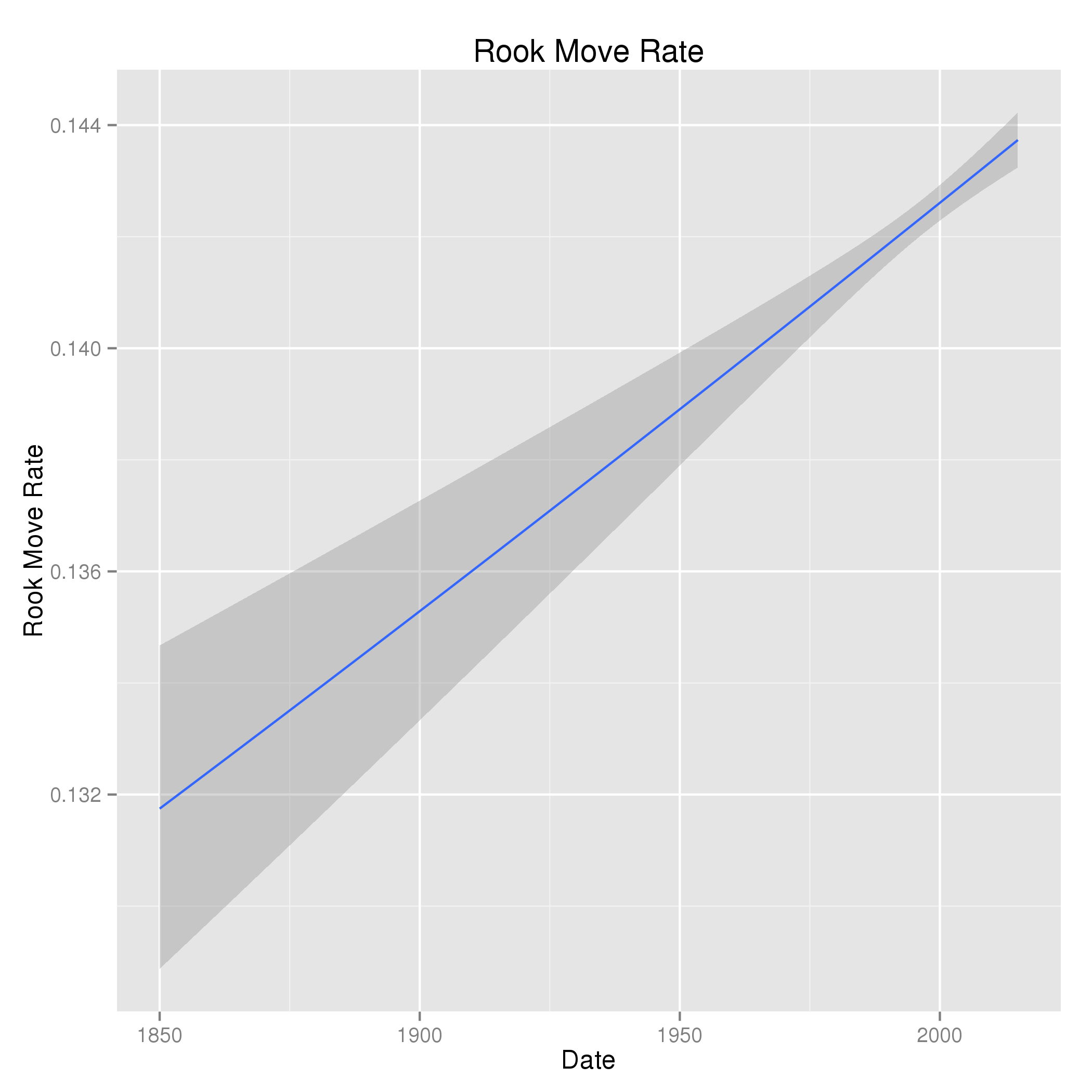}}
 \subfloat[]{\includegraphics[width=0.35\textwidth]{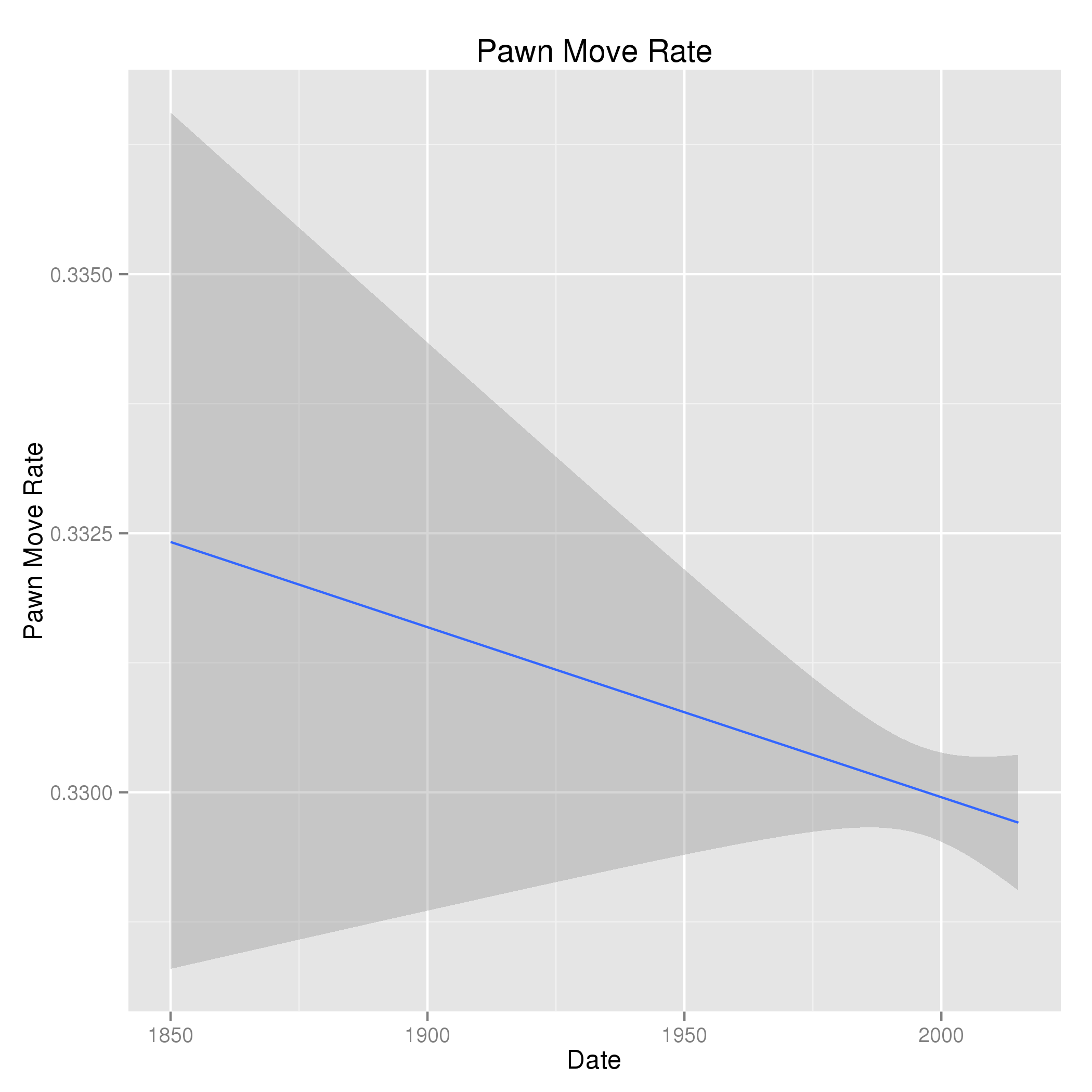}}
  	\hspace{3pt}
 \subfloat[]{\includegraphics[width=0.35\textwidth]{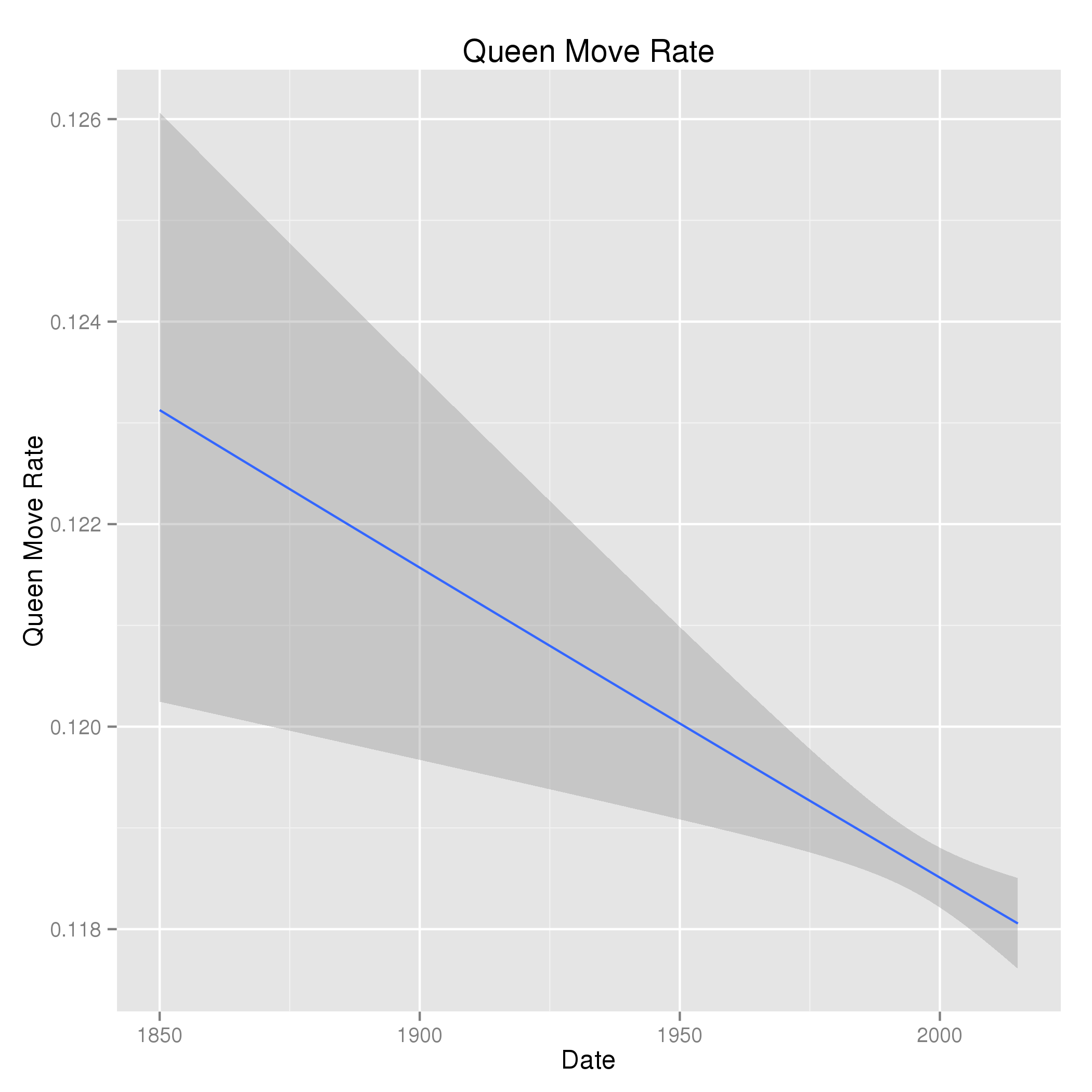}}
 \subfloat[]{\includegraphics[width=0.35\textwidth]{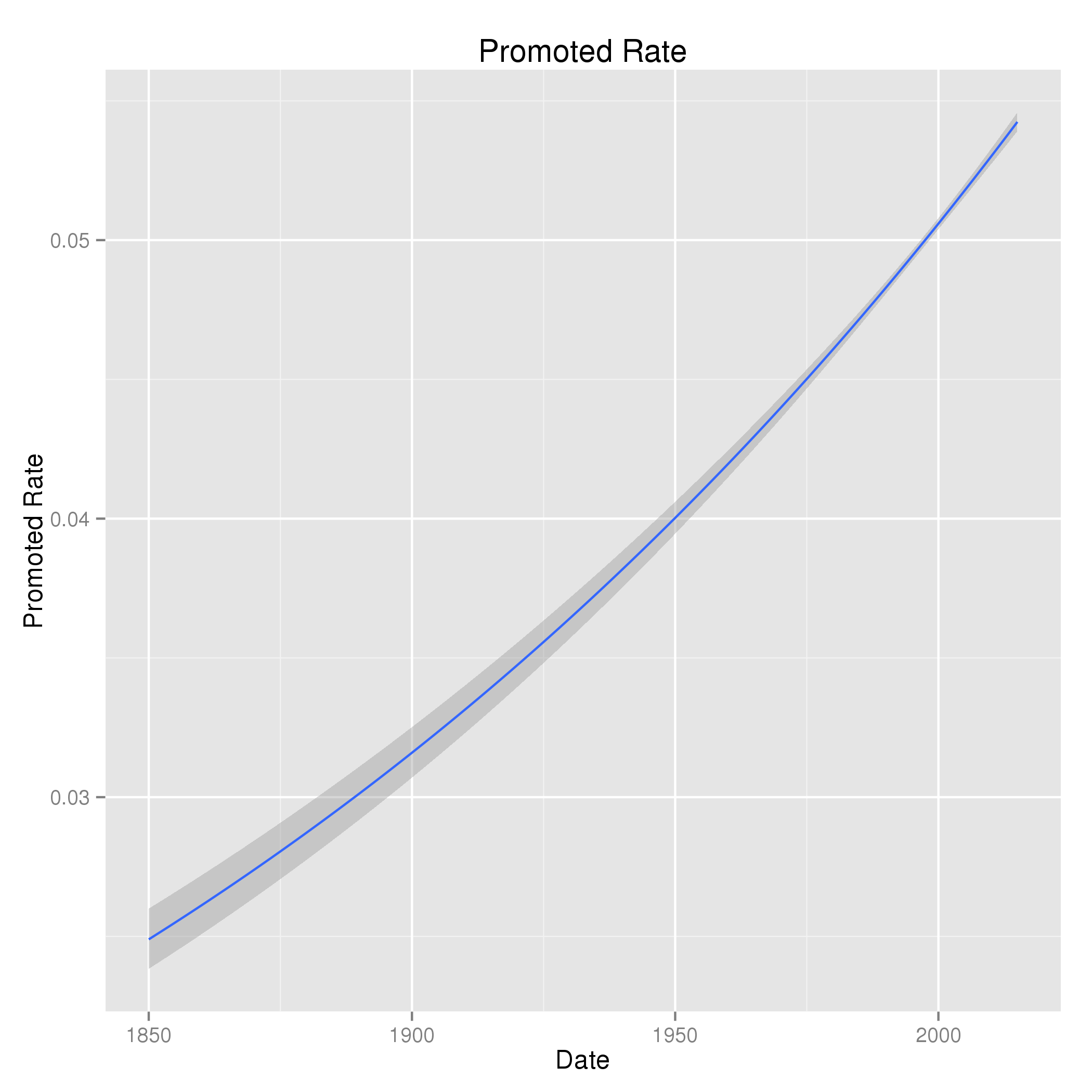}}
 \subfloat[]{\includegraphics[width=0.35\textwidth]{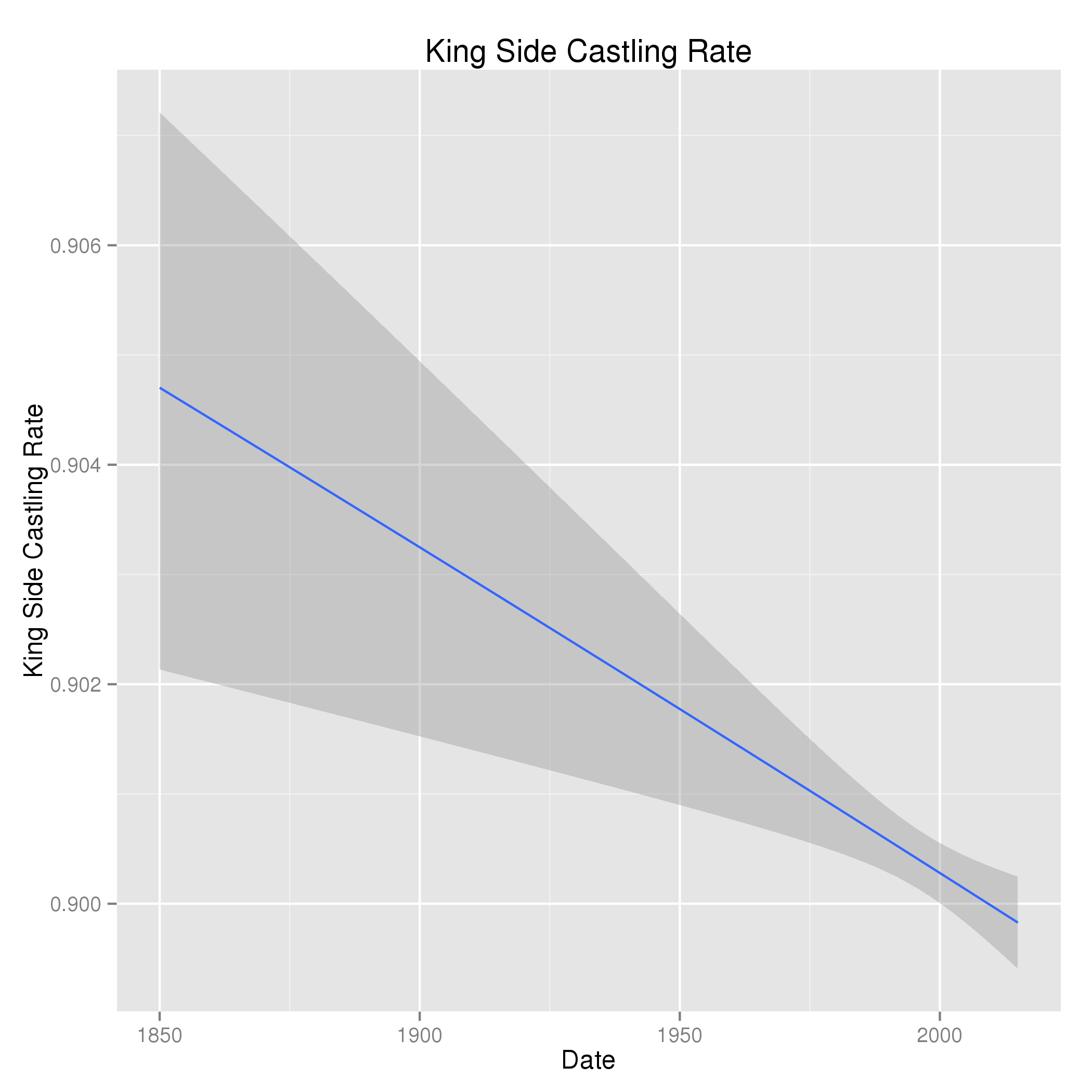}}
 	\hspace{3pt}
 \subfloat[]{\includegraphics[width=0.35\textwidth]{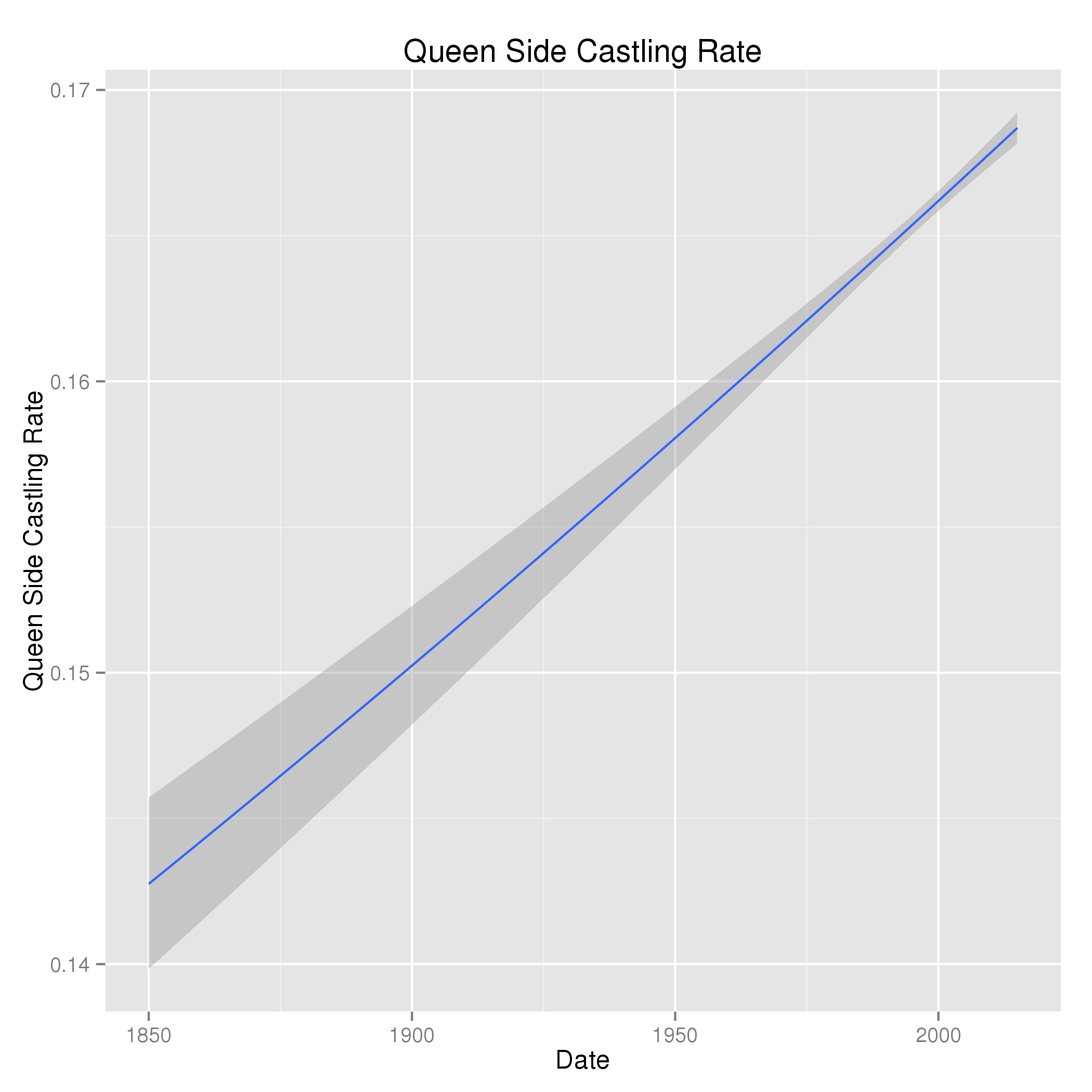}}
 \label{fig:rates}
 \caption{Pieces' rates}
\end{figure}



\end{document}